\documentclass[letterpaper, 10 pt, conference]{ieeeconf} 

\usepackage{times}
\usepackage{epsfig}
\usepackage{graphicx}
\usepackage{amsmath}
\usepackage{amssymb}
\usepackage{cite}
\usepackage{comment}
\usepackage{caption}
\usepackage{algorithm}
\usepackage{algpseudocode}
\usepackage{subfig}

\usepackage[breaklinks=true,bookmarks=false]{hyperref}

\begin{document}

\title{\LARGE \bf
Robotic Grasping using Deep Reinforcement Learning
}

\author{Shirin Joshi, Sulabh Kumra, Ferat Sahin\\
Rochester Institute of Technology, Rochester, NY, USA\\
{\tt\small \{saj8732, sk2881, fseee\}@rit.edu}
}

\maketitle

\begin{abstract}
In this work, we present a deep reinforcement learning based method to solve the problem of robotic grasping using visio-motor feedback. The use of a deep learning based approach reduces the complexity caused by the use of hand-designed features. Our method uses an off-policy reinforcement learning framework to learn the grasping policy. We use the double deep Q-learning framework along with a novel Grasp-Q-Network to output grasp probabilities used to learn grasps that maximize the pick success. We propose a visual servoing mechanism that uses a multi-view camera setup that observes the scene which contains the objects of interest. We performed experiments using a Baxter Gazebo simulated environment as well as on the actual robot. The results show that our proposed method outperforms the baseline Q-learning framework and increases grasping accuracy by adapting a multi-view model in comparison to a single-view model.

\end{abstract}

\section{Introduction}
Even though a lot of research has been done on robotic grasping, in real-world scenarios the robot cannot obtain a 100\% success rate in grasping a variety of objects. This is due to inaccuracies in the potential grasp detection and ultimately leads to the problem of not being able to select the best possible grasp for an object. A lot of recent work has addressed this by converting it into a detection problem which works on visual aspects of the image to infer the location where the robotic gripper needs to be placed.  
As compared to the use of hand-designed features used in the past which increases the time complexity, recent work focuses on the use of deep learning techniques that gives much better performance in the field of visual perception, audio recognition, and natural language processing. 

Grasping is primarily a detection problem and since a majority of the work in deep learning has been applied for recognition problems, previous applications of deep learning for detection have been specifically used for face detection, text detection, etc. This paper focuses on the use of a deep reinforcement learning framework to resolve the challenge of grasping an object with the help of a two-finger robotic gripper. The use of reinforcement learning in grasping enables the robot to learn to pick and place an object on its own without training it on a large dataset thereby eliminating the need for a dataset. Thus, the objective here is to not only find a feasible grasp but also to find the optimum grasp for an object that maximizes the chance of efficiently grasping it.

\begin{figure}
  \centering
    \includegraphics[width=0.48\textwidth]{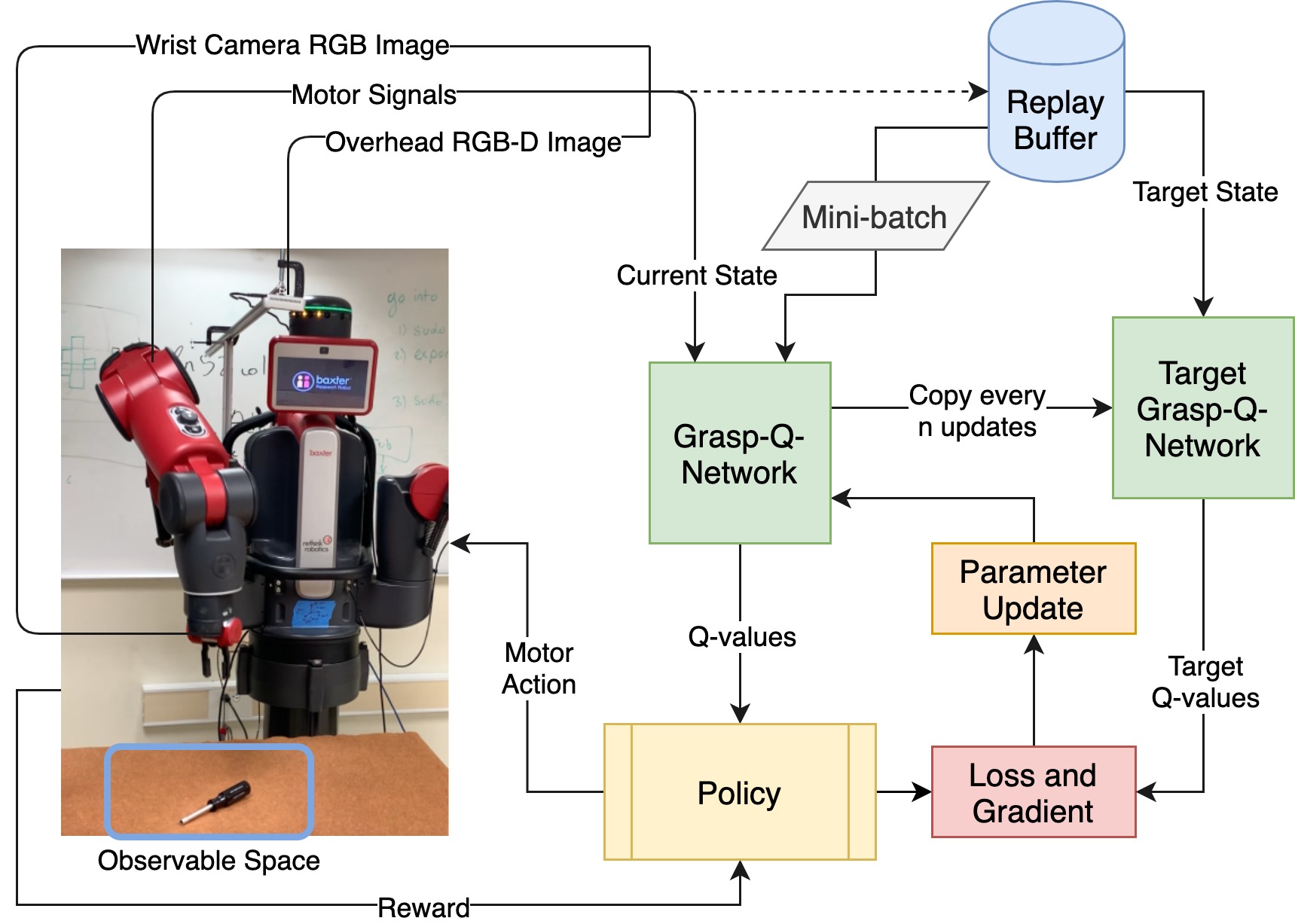}
    \caption{Overview of the proposed system}
    \label{fig: overview}
\end{figure}

Precisely modeling the physical environment every single time is practically impossible and signifies a lack of certainties. In this work, we study a grasping strategy through a deep reinforcement learning framework characterized by Grasp-Q-Network and continuous visual feedback. For this, a deep Q-network is trained, and the given task is assigned with rewards. Further, we improve the performance of this network by adapting an off-policy reinforcement learning framework. The proposed framework reduces the time and computation required for training the network and uses a small set of objects to train the robot instead of using a large database in order to learn the best policy to detect grasps for different objects and increases the number of successful grasps which lead to optimal outcomes.

The overview of our proposed system is shown in fig. \ref{fig: overview}. The agent observes the environment which contains an object to be grasped. Using the observation from the overhead camera and the wrist camera along with its current motor state, the Grasp-Q-Network outputs Q-values that are used by the agent to take action. The agent then receives a reward based on a policy. After each iteration, the loss function calculates the loss between the target Grasp-Q-Network and the Grasp-Q-Network and updates the parameters of the network. 

The key contributions of this work can be summarized as:
\begin{itemize}
    \item A novel deep reinforcement learning framework for learning robust grasps using multiple cameras and motor inputs. 
    \item A novel visual servoing mechanism to produce successful grasps using continuous visual feedback. 
    \item Evaluation of the performance of our method in a simulated as well as the real-world environment. We demonstrate that our novel double deep Q-learning network outperforms the proposed Deep Q-learning and the vanilla Q-learning framework by maximizing grasp success.
    \item Analysis of the performance of our method on different objects using single-view and multi-view camera setup. 
\end{itemize}


\section{Related Work}

{\bf Robotic Grasping:} There has been a lot of research in the field of robotic grasping to improve potential grasps of novel objects.  \cite{lenz2013deep,levine2018learning,Saxena2008RoboticGO,DBLP:journals/corr/KumraK16}.
Not only is the pose and configuration of the robotic arm and gripper important in grasp detection, but also the shape and physical properties of the object to be grasped are essential in determining how an object should be gripped. An acute understanding of the target object's 3D model is crucial for a swift pick and place action using a robotic manipulator. This can be achieved using 3D sensing cameras and sensors that give an almost good perception of the object in the real world. Grasping methods typically fall into two categories, particularly: Analytical methods and Empirical or data-driven methods \cite{6672028}.

The use of RGB-D cameras for object recognition, detection, and mapping has shown to improve the ratio of successful grasps, as seen from \cite{jiang2011efficient}. The use of RGB-D imagery is significantly useful since the depth information from the RGB-D images is utilized for mapping the 3D real-world co-ordinates to valuable information in the 2D frame. The previous work in the field of robotics focuses solely on these images obtained from a depth camera for grasp detection and object recognition \cite{lenz2013deep}.

{\bf Deep learning:} A lot of advancements have been made in recent years with the use of vision-based techniques in robotic grasping using deep learning \cite{lenz2013deep,levine2018learning,6739623,DBLP:journals/corr/abs-1802-10264,DBLP:journals/corr/KumraK16, mahler2019learning}. A majority of the work in deep learning is associated with classification, while only a few have used it for detection \cite{lenz2013deep, DBLP:journals/corr/KumraK16}. All of these approaches use a bounding box that contains the observed object, and the bounding box is similar for each valid object detected. However, for robotic grasping, there may be several methods to grasp an object.  But it is essential to pick the one with the highest grasp success or with the most stable grasp, thus relying on machine learning techniques to find the best possible grasp. The use of convolutional neural networks is a popular technique used for learning features and visual models that uses a sliding window detection approach, as illustrated by \cite{lenz2013deep}. However, this technique is slow for a robot in a real-world situation where it may be required to take fast actions. \cite{DBLP:journals/corr/KumraK16} worked on improving this by passing an entire image through the network rather than using small patches to detect potential grasps.

{\bf Reinforcement learning:} In reinforcement learning, the learner is not informed which action to take but instead, it should decide which action will yield the most reward by trial and error. In most cases, the actions not only affect the subsequent reward but the next action thereby affecting all the subsequent rewards \cite{sutton2018reinforcement}. Thus, trial and error search and delayed reward are the two most prominent features of reinforcement learning. Several others claim to use rules for grasping that is based on the research that portrays human actions for grasping and manipulating objects. \cite{549169} presents early work in using reinforcement learning for robotic grasps in which the learning approach is adopted from human grasping an object. Three layers of functional modules that enable learning from a finite set of data along with maintaining a good generalization \cite{1307456,10.1007/11552246_35} have shown to have successful implementations of reinforcement learning in the past. 

However, due to complex issues such as memory complexity, sample complexity, as well as computational complexity, the user has to rely on deep learning networks. These networks use function approximations and representation learning properties to overcome the problems of using algorithms that require very high computational power and fast processing. \cite{5979644} discusses the use of a reinforcement learning algorithm, Policy Improvement with Path Integrals (PI$^{2}$) that can be used when the state estimation is uncertain and this approach does not require a specific model and is thus model-free. \cite{6707053} discusses how a system can learn to reliably perform its task in a short amount of time by implementing a reinforcement learning strategy with a minimum amount of information provided for a given task of picking an object. Deep learning has enabled reinforcement learning to be used for decision-making problems such as settings with large dimensional state and action spaces that were once unmanageable. In recent years, a number of off-policy reinforcement learning techniques have been implemented. For instance, \cite{DBLP:journals/corr/MnihKSGAWR13} uses deep reinforcement learning for solving Atari games, \cite{tai2017virtual} uses a model-free algorithm based on the deterministic policy gradient to solve problems in continuous action domain.

The current research on deep Q-network (DQN) shows how deep reinforcement learning can be applied for designing closed-loop grasping strategies. \cite{kalashnikov2018scalable} demonstrates this by proposing a Q-function optimization technique to provide a scalable approach for vision-based robotic manipulation applications. \cite{zeng2018learning} uses deep Q-learning for learning grasping strategies, along with pushing for applications comprising tightly packed spaces and cluttered environments. Our proposed work is based on a similar approach that uses an adaptation of the deep reinforcement learning technique for detecting robot grasps.


\begin{figure*}
    \centering
    \includegraphics[width=0.9\textwidth]{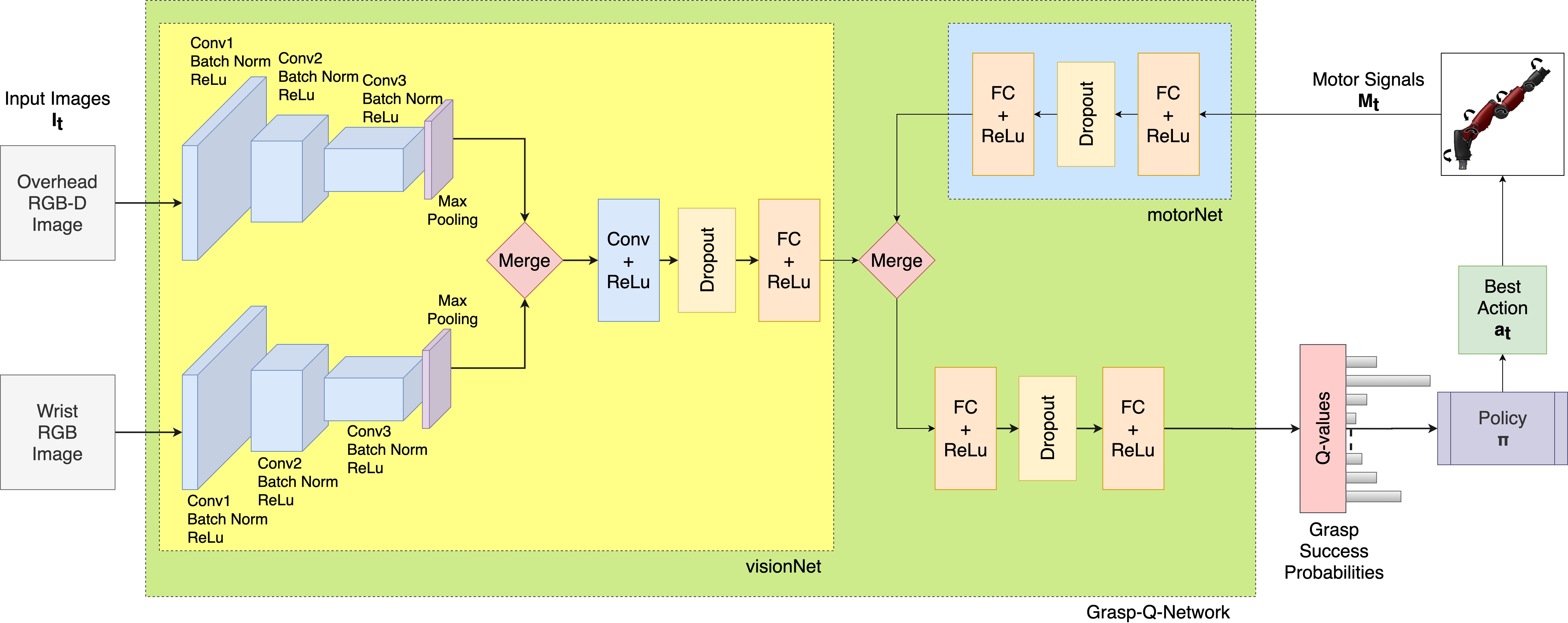}
    \caption{The architecture of proposed Grasp-Q-Network. The input RGB-D image from the overhead camera $\mathbf { I }_{ t }^{o}$ along with the wrist RGB camera $\mathbf { I}_{ t }^{w}$ are individually fed to a 7 $\times$ 7 convolution with stride 2, followed by Batch Normalization. This is followed by a 5 $\times$5 convolutional followed by a 3$\times$3 convolution followed by Batch Normalization and  Max-pooling. The output features are then concatenated and fed to a convolutional layer followed by a fully connected layer. The servo motor command $\mathbf { M } _ { t }$ is processed by two fully connected layers. The result is then processed by two fully connected layers, after which the network outputs the probability of grasp success using a softmax function.}
    \label{drl}
\end{figure*} 
  

 \section{Problem Formulation}
We define the problem of robotic grasping as a Markov Decision Process (MDP) where at any given state $ { s } _ { t } \in \mathcal{S}$ at time ${t}$, the agent (i.e. robot) makes an observation ${o}_{t} \in \mathcal{O}$ of the environment and executes an action ${ a } _ { t } \in \mathcal{A}$ based on policy $\pi({ s } _ { t })$ and receives an immediate reward ${r}_{t}$ based on the reward function $\mathcal{R} \left( { s } _ { t },  { a } _ { t } \right)$. The goal of the agent is to find an optimal policy $\mathcal{\pi^{*}}$ in order to maximize the expected sum of discounted future rewards i.e. $\gamma$-discounted sum on all future returns from time $t$ to $\infty$.  

In our work, the observation ${o}_{t}$ comprises of the RGB-D image captured from the overhead depth camera and the RGB image taken from the wrist camera along with the joint angles of the robot's arm observed at time $t$.


\section{Proposed Approach}

We propose a novel deep reinforcement learning based technique for vision-based robotic grasping. The proposed framework consists of a novel double deep Q-learning based architecture. The architecture consists of observed input taken from the overhead camera and the wrist camera along with the current motor positions, fed into the Grasp-Q-Network, that returns the grasp success probabilities i.e. the Q values for all possible actions that the agent can take. These Q-values are then used to select the best action $\mathbf{a}_{t}$ based on the $\epsilon$-greedy policy to find the optimal policy $\pi^{*}$. 



\subsection{Overview}

The overview of the proposed system is shown fig. 2. The method presented involves end-to-end training of the Grasp-Q-Network based on a visual servoing mechanism that performs continuous servoing to adjust the motor commands of the robot by observing the current state of the environment to produce a successful grasp. The motor actions are in the robot frame, therefore there is no requirement for the camera to be calibrated precisely with respect to the end effector.


The model uses visual indicators to establish the connection between the graspable object and the robot end-effector. A novel Deep Convolutional Neural Network is used that decides what motion of robot arm and the gripper can produce an effective grasp. Each potential grasp consists of time step T. For each of this time steps the robot records current images $ { I }_{ t }^{o}$ and $ { I }_{ t }^{w}$ from the overhead and wrist camera respectively and the current motor positions ${ M } _ { t }$. Using the policy $\pi(\mathbf { s } _ { t })$, the agent determines the next action, i.e. in which direction to move the gripper. The agent then evaluates the action for each time step ${t}$ and produces a reward ${ r } _ { t }$ using the reward function $ \mathcal{R} \left( \mathbf { s } _ { t } , \mathbf { a } _ { t } \right)$. Each of these time steps T results in training samples $\mathcal{T}$ given by the equation:
\begin{equation}
    \mathcal{T} = \left( {I}_{t}^{o} ,  {I}_{t}^{w}, {M}_{t}, {a}_{t}, {r}_{t} \right)
\end{equation}
Each time step involves the observed sample images, the motor position information, the action taken, and the reward received. The action space $\mathcal{A}$, is defined as a vector of motor actions comprising of end-effector displacement along the three axes, rotation along the z-axis, and the gripper action that involves closing the gripper which terminates the episode.

The Grasp-Q-Network comprises of two parts: \textit{visionNet} and \textit{motorNet} as shown in fig.\ref{drl}. The Grasp-Q-Network outputs probabilities as Q-values for each of the actions in action space $\mathcal{A}$, which are used to select the best action ${a}_{t}$. The Grasp-Q-Network can be considered as Q-function approximator for the learning policy described by the visual servoing mechanism $\lambda \left({ o } _ { t } \right)$. The repeated utilization of the target GQN that is defined by $\psi^{\prime} \left( { I } _ { t } ,  { M } _ { t } \right)$ to get more data to refit the GQN $ \psi \left( { I } _ { t } , { M } _ { t } \right)$ can be regarded as fitted Q iteration.

\begin{algorithm}
\caption{Proposed Visual Servoing Mechanism}
\begin{algorithmic}[1]
\Procedure{Servoing}{$\lambda \left({ o } _ { t } \right)$}

\State \text{Given current images $ { I }_{ t }^{o}$ and $ { I }_{ t }^{w}$}
\State \text{Given current motor position ${M}_{t}$}

\State \emph{top}:
\State \text{Initialize time step ${t}$ and reward ${r}$}

\State \emph{loop}:

\State \text{Infer ${a}_{t}$ using network $ \psi \left({ I } _ { t }, { M } _ { t } \right)$ and policy $\pi^{*}$}
\State \text{Execute action ${a}_{t}$}

\If{${a}_{t}$ is not executed} 
    \State $r \gets r - 1$
    \State \textbf{goto} \emph{top}
\EndIf

\If{$\textit{result} = \textit{successful pick}$}
    \State{$r \gets r + 10$}
    \State \textbf{goto} \emph{top}
\ElsIf{$\textit{result} = \textit{partial pick}$}
    \State{$r \gets r + 1$}
    \State \textbf{goto} \emph{top}
\Else{}
    \State{$r \gets r - 0.025$}
\EndIf

\State \text{$t \gets t + 1$}
\State \textbf{goto} \emph{loop}.

\EndProcedure
\end{algorithmic}
\end{algorithm}


\subsection{DQN Algorithm}
A deep Q network (DQN) is a multiple layer neural network that returns a vector of action values $Q \mathbf ( s, a \mid \boldsymbol { \phi } )$ for a given state ${s}_{t}$ and sequence of observations $\mathcal {O}$, where $\phi$ represents the parameters of the network. The main aim of the policy network is to continuously train $\phi$ to attain the optimal policy $\pi^{*}$. Two important aspects of the DQN algorithm, as proposed by \cite{mnih2015human}, are the use of a target network and the use of experience replay.

In our work, the target network $\psi^{\prime}$, which is parameterized by $\boldsymbol {\phi}^{\prime}$, is same as the GQN $\psi$ except that its parameters are copied every ${N}$ time steps from the GQN so that then $\boldsymbol { \phi} _ { t=N } ^ {\prime} = { \phi } _ { t=N }$ and kept fixed on all other steps. The target used by DQN is then:
\begin{equation}
Y _ { t } ^ { \mathrm { DQN } } \equiv r _ {t} + \gamma \max _ { a } Q \left( s _ { t + 1 } , a \mid \boldsymbol { \phi } _ { t } ^ { \prime } \right)
\end{equation}
\noindent where, $\gamma \in [0, 1)$ is the discount factor. A value closer to $0$ indicates that the agent will choose actions based only on the current reward and values approaching $1$ indicates that the agent will choose actions based on the long-term reward.

\subsection{Training}
Two separate value functions are trained in a mutually symmetric fashion by designating experiences arbitrarily to update one of the two value functions, that results in two set of weights $\boldsymbol { \phi } \text { and } \boldsymbol { \phi } ^ { \prime }$. One set of weights is used to learn the greedy policy, and the other is used to determine its value for each update. To compare with Q-learning target, which is defined as:
\begin{equation}
Y_{ t } ^ { \mathrm { Q } } \equiv r_{t} + \gamma \max _ { a } Q \left( s_{ t + 1 } , a_{t+1} \mid { \phi } \right)
\end{equation}

\noindent where, $r_{t}$ is immediate reward, and $s_{ t + 1 }$  is the resulting state, and $Q \left( s , a \mid { \phi } \right)$ is the parameterized value function. The target value in double deep Q-learning can be written as:
\begin{equation}
 Y _ { t } ^ { DDQN } = r_{t} + \gamma Q \left( s _ { t + 1 } , \operatorname { argmax } Q \left( s _ { t + 1 } , a _ { t + 1 } \mid { \phi } \right) \mid { \phi } ^ { \prime } \right) 
\end{equation}

This solves the overestimation issue faced with Q-learning and outperforms the existing Deep Q-learning algorithm. Instead of the value obtained from the target network  ${\psi}_{\phi}^ {\prime}$, the action that amplifies the current network ${\psi}_{\phi}$ is used by the max operator in double Q-learning.

Thus, double DQN can be seen as a policy network ${\psi}$, parameterized by ${\phi}$, which is continually being trained to approximate the optimal policy. Mathematically, it uses the Bellman equation to minimize the loss function $\mathcal{L}(\phi)$ as below:
\begin{equation}
 \mathcal{L}(\phi) = \mathbb{E} \left[ \left( Q \left( s _ { t} , a _ { t} \mid { \phi } \right) - Y _ { t } ^ { DDQN }\right) ^ 2 \right]
\end{equation}

\subsection{Reward Function}
The reward for a successful grasp is given by $ \mathcal{R} \left( \mathbf { s } _ { t } , \mathbf { a } _ { t } \right) = 10$ . A grasp is defined as successful if the gripper is holding an object and is above a certain height at the end of the episode. A partial reward equal to 1 is given if the end effector comes in contact with the object. A reward of -1 is given if no action is taken. Moreover, a small penalty of -0.025 is given, for all time steps prior to termination, to encourage the robot to grasp more quickly. 


\section{Experiments}

\begin{figure*}
    \centering
    \includegraphics[width=0.9\textwidth]{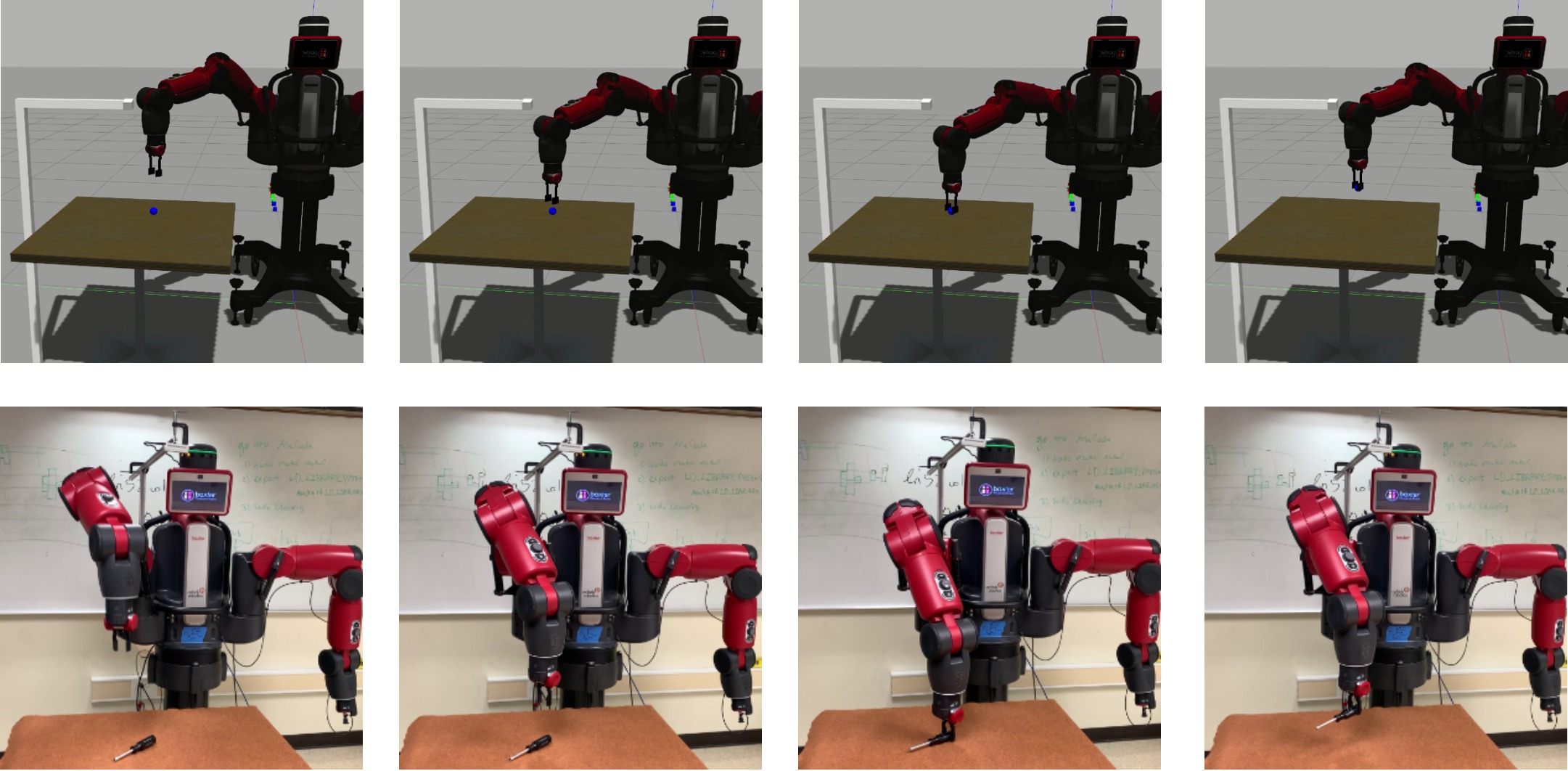}
    \caption{Baxter learning to grasp objects using the proposed deep reinforcement learning based visual servoing algorithm in simulation (top row) and real-world (bottom row)}
    \label{fig: baxter-picking}
\end{figure*}

To evaluate our approach, we perform our experiments using our proposed deep Q-learning and double deep Q-learning algorithms. We use the vanilla Q-learning algorithm as a baseline. The experiments were performed in simulation as well as in the real-world on the 7-DOF Baxter robot. Further, the experiments were carried out using a single as well as multiple camera setups. 

\subsection{Setup}
\begin{figure}
    \centering
    \includegraphics[width=8.5cm, height=10cm,keepaspectratio]{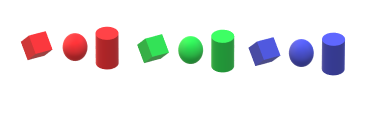}
    \caption{3D model of objects used for training the robot}
    \label{fig:model_of_objects}
\end{figure}

The simulation setup involves interfacing the gazebo simulator in Robot Operating System (ROS) with Atari DQN that provides a custom environment for simulating our proposed method. Although the proposed approach is robot agnostic, we use a Baxter robot with a parallel plate gripper as our robot platform. We create a simulation environment with a table placed in front of the robot and a camera mounted on the right side of the table. An object is randomly spawned on the table at the start of each episode. The objects consist of a cube, a sphere, and a cylinder as seen from fig. \ref{fig:model_of_objects}.

The real-world setup consisted of the Baxter robot, Intel RealSense D435 depth camera used as an overhead camera, and a wrist camera similar to the simulation setup. The overhead camera position was adjusted in real-world implementation to avoid collision with the robot.

Fig. \ref{fig: baxter-picking} top row demonstrates an example run for simulation of the Baxter robot learning to pick objects in the Gazebo environment. The bottom row shows the robot approaching an object and the successful completion of the task of picking an object in the real-world setting.  

\subsection{Training}
For training, at the start and reset, a colored sphere, cylinder, or cube is spawned/placed at a random orientation in front of the Baxter robot. The robot then tries to pick up the object by maneuvering its arm around the object. In our work, the arm movement is limited to a rotation around the wrist and shoulder along with the ability to extend the reach while the gripper is facing downwards. At the end of the pickup action, the success of the task is measured by checking the gripper feedback. The gripper close status where the gripper is not fully closed indicates that an object is grasped and gripper fully closed indicates that the object is not grasped. A partial reward is given if the robot comes in contact with the object. The environment is reset at the termination of the episode. 

The arm movement is based on the visual servoing mechanism between the overhead camera and the right-hand wrist camera, presenting the current observation of the scene with the target object in comparison to the current position of the gripper. The image data consists of an image from the overhead camera and an image from the wrist camera. The image data, along with motor position information, is sent from the robot to the Grasp-Q-Network, which processes the information obtained from the image, and in turn, sends servo commands back to the robot. Objects are randomized in the start position and can appear randomly in any position on the table within the robot's reach. The gripper closes when it moves to the position vector obtained from the Grasp-Q-Network to pick the object, and the episode ends after the object is successfully picked up from its location or the maximum number of steps per episode is reached.

\subsection{Results}
\begin{figure}
    \centering
    \includegraphics[width=.9\linewidth]{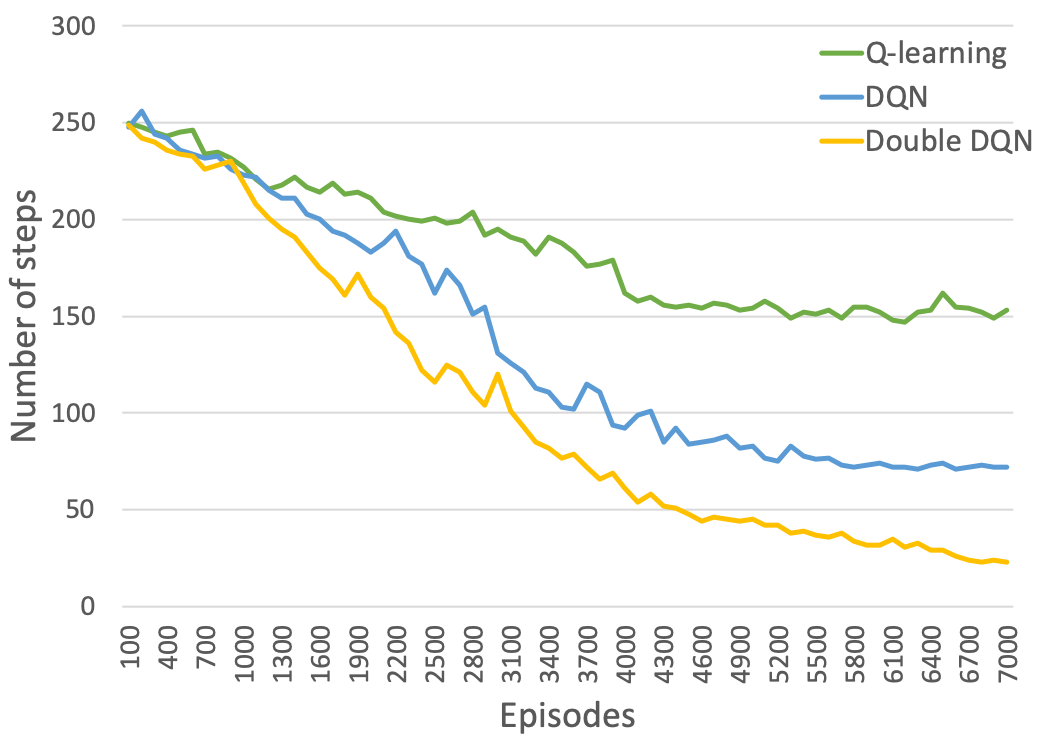}
    \caption{Learning curves of Q-learning, DQN and double-DQN on the robotic grasping task}
    \label{fig:results}
\end{figure}

The resulting plot, as seen in fig. \ref{fig:results} shows that our double-DQN based algorithm performs better than the vanilla Q-learning algorithm. The results demonstrate that DQN as well as double-DQN have comparatively higher completion and success rates than the baseline Q-learning framework. The experiments were performed for a total of 7000 episodes, and it can be observed that the number of steps required for the successful completion of the task is least for the double deep Q-learning method followed by the deep Q-learning method. Further, our experiment shows that this method can be adapted for the real robot by adjusting the configurations of the robot software environment.

\begin{figure}
    \centering
    \includegraphics[width=.9\linewidth]{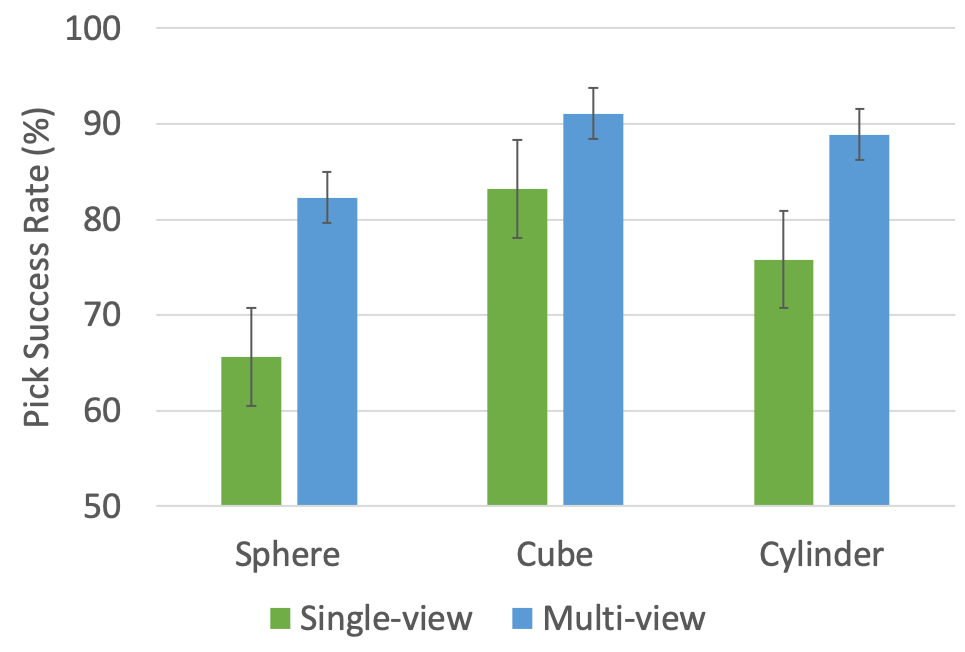}
    \caption{Comparison of performance of single and multi-view models for the robotic grasping task}
    \label{fig:single-multi-results}
\end{figure}

The use of a single camera may often consist of occlusions due to robot arm. For this reason, we also compare the results of single and multi-view camera setups to test the performance of the robot for picking three objects: sphere, cube, and cylinder, as shown in fig. \ref{fig:single-multi-results}. The single camera setup involves using a single camera input, and multiple camera setup incorporates an overhead camera and a wrist camera of the robot arm. The results demonstrate that the multi-view camera model works better in comparison to the single-view model for all three objects. An interesting observation to note is that both the single-view and multi-view models work best on the cube with an accuracy of 91.1\% on a multi-view model, and 83.2\% on a single view model as compared to the sphere and the cylinder, and have the lowest success rate for the sphere.

\section{Conclusions}
A method for learning robust grasps is presented using a deep reinforcement learning framework that consists of a Grasp-Q-Network which produces grasp probabilities and a visual servoing mechanism that performs continuous servoing to adjust the servo motor commands of the robot. In contrast to most grasping and visual servoing methods, this method does not require training on large sets of data since it does not use any dataset for learning to pick up the objects.
The experimental results show that this method works in simulations as well as on the real robot. This method also uses continuous feedback to correct inaccuracies and adjust the gripper to the movement of the object in the scene. Our results on using multi-view models show that multi-view camera setup works better in comparison to single view camera setup and also has a higher success rate for multiple cameras. 
Further, the proposed novel Grasp-Q-Network has been used to learn a reinforcement learning policy that can learn to grasp objects from the feedback obtained by the visual servoing mechanism. Finally, the results obtained show that by adapting an off-policy reinforcement learning method, the robot performs better at learning the task of grasping the objects as compared to the vanilla Q-learning algorithm, and reduces the overestimation problem that was observed using the deep Q-learning algorithm.  




{\small
\bibliographystyle{ieeetr}
\bibliography{egbib}
}


\end{document}